\documentclass[10pt,twocolumn,twoside]{IEEEtran}
\IEEEoverridecommandlockouts

\usepackage{cite}
\usepackage{amsmath,amssymb,amsfonts}
\usepackage{algorithmic}
\usepackage{graphicx}
\usepackage{textcomp}
\usepackage{balance}
\usepackage{xcolor}
\usepackage[utf8]{inputenc}

\usepackage{amsthm}
\usepackage{enumitem}

\usepackage{amsmath,amsfonts,mathtools,algorithm,booktabs,bm,hyperref,xcolor}

\newcommand{\x}{\mathbf{x}}
\newcommand{\y}{\mathbf{y}}
\newcommand{\param}{\theta}
\newcommand{\threshold}{\epsilon}
\newcommand{\PrimalSet}{\mathcal{X}_{\param}}
\newcommand{\DualSet}{\mathcal{Y}_{\param}}
\newcommand{\ParamSet}{\Theta}  % set of valid parameters
\newcommand{\pobj}{\phi}
\newcommand{\dobj}{\psi}
\newcommand{\popt}{\phi^{\star}}
\newcommand{\dopt}{\psi^{\star}}
\newcommand{\xopt}{\x^{\star}}
\newcommand{\yopt}{\y^{\star}}
\newcommand{\xpred}{\hat{\x}}
\newcommand{\ypred}{\hat{\y}}

\newcommand{\Gap}{\Gamma}
\newcommand{\Dataset}{\mathcal{D}}

\newcommand{\pweights}{\alpha}
\newcommand{\pproxy}{p_\pweights}
\newcommand{\dweights}{\beta}
\newcommand{\dproxy}{d_\dweights}

\newcommand{\PG}{\mathbf{p}^{\text{g}}} % active power generation
\newcommand{\PD}{{p}^{\text{d}}} % active load
\newcommand{\PF}{\mathbf{p}^{\text{f}}} % active flow
\newcommand{\PFhat}{\hat{\mathbf{p}}^{\text{f}}} % active power generation
\newcommand{\PGhat}{\hat{\mathbf{p}}^{\text{g}}} % active power generation
\newcommand{\PGtil}{\tilde{\mathbf{p}}^{\text{g}}} % active power generation
\newcommand{\PFtil}{\tilde{\mathbf{p}}^{\text{f}}} % active power generation

\newfloat{model}{thp}{lop}\floatname{model}{Model} % algorithm-style

\def\BibTeX{{\rm B\kern-.05em{\sc i\kern-.025em b}\kern-.08em
    T\kern-.1667em\lower.7ex\hbox{E}\kern-.125emX}}
\usepackage{lipsum}

\title{\LARGE \bf 
Self-Certifying Primal-Dual Optimization Proxies\\ for Large-Scale Batch Economic Dispatch}

\author{%
Michael Klamkin$^{\dagger\ddagger}$, Mathieu Tanneau$^\dagger$, Pascal Van Hentenryck$^\dagger$% <-this % stops a space
\vspace{-0.5cm}
 \thanks{ %
$^\dagger$Affiliated with NSF AI Institute for Advances in Optimization, Georgia Institute of Technology, Atlanta, Georgia, United States.}
\thanks{$^\ddagger$Corresponding Author. Email: {\tt klam@gatech.edu}.}
}

\begin{document}
\begingroup
\allowdisplaybreaks

\maketitle

\begin{abstract}
Recent research has shown that optimization proxies can be trained to high fidelity,
achieving average optimality gaps under 1\% for large-scale problems.
However, worst-case analyses show that there exist in-distribution queries
that result in orders of magnitude higher optimality gap,
making it difficult to trust the predictions in practice.
This paper aims at striking a balance between classical solvers and
optimization proxies in order to enable \textit{trustworthy deployments
with interpretable speed-optimality tradeoffs based on a user-defined optimality threshold}.
To this end, the paper proposes a hybrid solver that leverages duality theory
to efficiently bound the optimality gap of predictions, falling back to a
classical solver for queries where optimality cannot be certified. 
To improve the achieved speedup of the hybrid solver, the paper proposes
an alternative training procedure that combines the primal and dual proxy training.
Experiments on large-scale transmission systems show that the hybrid solver is highly scalable. 
The proposed hybrid solver achieves speedups of over 1000x compared to a parallelized
simplex-based solver while guaranteeing a maximum optimality gap of 2\%.
\end{abstract}


\section{Introduction}
\label{sec:introduction}

In order to plan and operate modern power grids, Transmission Systems Operators (TSOs)
regularly have to solve large numbers of parametric optimization problems, e.g.,
multi-year simulations for transmission and capacity planning \cite{gu2011market},
or Monte-Carlo simulations for short-term risk analysis \cite{stover2023reliability,chen2024real}.
Parametric optimization problems also appear in stochastic optimization formulations,
which have received growing interest as renewable generation and
distributed energy resources are integrated into the grid \cite{knueven2023stochastic,brahmbhatt2025benders}.

The computational cost of solving large batches of similar instances (in the tens or hundreds of thousands) has fueled significant interest in \textit{optimization proxies}, Machine Learning-based (ML) surrogate models that emulate the behavior of optimization solvers.
Once trained, optimization proxies can produce close-to-optimal solutions to large-scale problems
in milliseconds \cite{khaloie2024review,van2025optimization}, bringing performance guarantees to proxy-based market clearing and risk analysis \cite{chen2024real}.
Nevertheless, although proxies enjoy inference speeds that are orders of magnitude faster than classical optimization algorithms, their lack of strong worst-case performance guarantees has hindered their widespread adoption in real-life settings.

To address this limitation, the paper proposes a hybrid solver framework that, for the first time,
combines the speed of data-driven proxies with the worst-case guarantees of optimization solvers.
A core component of this framework is to jointly learn primal \emph{and} dual feasible solutions,
thus providing a self-contained mathematical certificate of performance in the form of a provable duality gap.
This unique self-certifying capability enables the systematic detection of poor predictions \emph{online},
without ground truth information. Namely, when the duality gap of a predicted primal-dual solution
exceeds a user-prescribed optimality tolerance, the corresponding instance is solved
exactly using a classical optimization solver. This strategy ensures worst-case optimality guarantees
while providing a large overall speedup since, in practice, only a small fraction of samples exhibit poor performance.
In addition, it provides an interpretable way to tradeoff optimality and speedup. Such optimality guarantees are crucial in the context of  market clearing with performance requirements and real-time risk analysis based on proxies \cite{chen2024real}, ensuring that the simulated system states are consistent with actual operations.

\subsection{Related Works}

\subsubsection{Optimization Proxies}

Recent research has explored how to design \textit{feasible} optimization proxies:
architectures that guarantee by construction the feasibility of the solution 
\cite{amos2017optnet,kim2022projection,chen2023end,tordesillas2023rayen,min2024hardnet,constante2025enforcing}.
In particular, \cite{chen2023end}
proposes a differentiable repair layer based on proportional response to efficiently produce 
primal-feasible solutions to the economic dispatch problem.
Nevertheless, despite substantial progress on the theory and practice of optimization proxies,
their widespread adoption remains hindered by their lack of worst-case performance guarantees.
Indeed, \cite{nnvopf} show that, while proxies may achieve low optimality gaps on average
(e.g. under 1\%), their worst-case performance can be orders of magnitude larger
(e.g. above 100\%), well beyond practically reasonable tolerances.

\subsubsection{Neural Network Verification}

In order to enable responsible adoption of proxies in practice, recent work has began to consider how to bound the sub-optimality of trained proxies by adapting techniques from neural network verification (NNV) \cite{liu2021algorithms}.
\cite{nellikkath2021physics} applies MIP reformulation techniques for verifying the worst-case infeasibility of ACOPF proxies. 
\cite{nellikkath2024scalable}  proposes primal acceleration techniques, using state-of-the-art NNV solvers \cite{alphabetacrown} for verifying DCOPF proxies. 
\cite{nnvopf} considers primal-feasible proxies for economic dispatch, proposing a first-order method to approximate the worst-case sub-optimality.

It is important to note that \cite{nnvopf,nellikkath2021physics,nellikkath2024scalable}
all show that worst-case performance in trained optimization proxies is often
orders of magnitude worse than average performance; in other words, optimization proxies are not immune to adversarial attacks. 
Thus, relying on neural network verification tools as a screening procedure for proxy adoption, besides being computationally expensive, leads to overly pessimistic conclusions. 

\subsubsection{ML-accelerated Optimization}

    Another line of work considers warm-starting schemes,
    where machine learning predictions are used to set the initialization of a classical solver. 
    Then, the overall scheme inherits both feasibility and optimality guarantees from the termination of the classical solver. 
    Prior works in this direction report speedups on the order of 2-25x \cite{xu2025explainable,dong2020smart,park2023compact,diehl2019warm}.
    This paper strikes a balance between warm-starting and directly using predictions; it proposes a scheme that allows to avoid the solver entirely for a majority of queries, unlocking further speedup on the order of 1000x.

\subsection{Contributions and Outline}

The paper contributions are summarized as follows:
\begin{enumerate}[labelindent=0pt,parsep=2pt, itemsep=0pt, topsep=2pt]
    \item It introduces a principled framework that fuses AI and optimization and
    provides, for the first time,  \textit{controllable }worst-case optimality guarantees without
    1) requiring any expensive offline verification step or
    2) imposing restrictions on the underlying architectures.
    Importantly, the optimality tolerance is prescribed by users
    rather than being derived from properties of the underlying models.
    \item It improves the training of primal-dual proxies by taking advantage of their self-certifying nature and uses a dedicated loss function.
    \item It presents a complete implementation of the proposed framework
    for the economic dispatch problem and conducts extensive numerical experiments
    on large-scale instances based on the European transmission system.
    \item {It demonstrates state-of-the-art performance for solving large batches
    of industry-scale economic dispatch problems. On the 9241\_pegase test case, the hybrid solver achieves up to 925x speedup over a parallelized simplex-based solver while guaranteeing worst-case optimality below 1\%, and over 1000x at 2\%.}
\end{enumerate}

\noindent
The rest of the paper is organized as follows:
Section \ref{sec:hybrid_solver} proposes the hybrid solver framework,
Section \ref{sec:primal_dual_training} describes the proposed training algorithm, 
Section \ref{sec:exp} presents the numerical experiments, and
Section \ref{sec:conclusion} concludes the paper.

\section{Self-Certifying Proxies}
\label{sec:hybrid_solver}

This section introduces the notations used throughout the paper by reviewing linear programming, duality, and optimization proxies. Then, it presents the hybrid solver, which exploits the self-certifying property of primal-dual feasible solutions.

    \subsection{Linear Programming and Duality}
    \label{sec:background_lp}

        For ease of reading, the proposed framework is stated for general linear programming (LP) problems.
        Namely, consider a parametric LP problem in primal-dual form
        \begin{subequations}
        \begin{align}
            \label{eq:background:LP:primal}
            P_{\param}: \quad &
                \min_{\x} \quad 
                \left\{ 
                    c_{\param}^{\top}\x
                \ \middle| \ 
                    A_{\param} \x \geq b_{\param}
                \right\},
            \\
            \label{eq:background:LP:dual}
            D_{\param}: \quad &
                \max_{\y \geq 0} \quad
                \left\{
                    b_{\param}^{\top}\y
                \ \middle| \ 
                    A^{\top}_{\param} \y = c_{\param}
                \right\},
        \end{align}
        \end{subequations}
        where $\param \, {\in} \, \mathbb{R}^{p}$ is the vector of parameters, which takes value in $\ParamSet \, {\subseteq} \, \mathbb{R}^{p}$, and $A_{\param} \, {\in} \, \mathbb{R}^{m \times n}$  $b_{\param} \, {\in} \, \mathbb{R}^{m}$ and $c_{\param} \, {\in} \, \mathbb{R}^{n}$  are the (parametric) left-hand side matrix, right-hand side, and objective, respectively.
        For $\param \, {\in} \, \ParamSet$, the primal and dual feasible sets are
        \begin{subequations}
        \begin{align}
            \label{eq:background:primal_feasible}
            \PrimalSet &= \{ \x \in \mathbb{R}^{n} \ | \ A_{\param} \x \geq b_{\param} \},\\
            \label{eq:background:dual_feasible}
            \DualSet &= \{ \y \in \mathbb{R}^{m} \ | \ A^{\top}_{\param} \y = c_{\param}, \y \geq 0 \}.
        \end{align}
        \end{subequations}
        The paper assumes that $\PrimalSet \, {\neq} \, \emptyset$ and $\DualSet \, {\neq} \, \emptyset, \forall \param \, {\in} \, \ParamSet$, i.e., the primal-dual feasible set is always non-empty.
        This mild technical assumption ensures that strong duality always holds.
        
        The primal (resp. dual) objective value of a primal-feasible (resp. dual-feasible) solution $\x \, {\in} \, \PrimalSet$ (resp. $\y \, {\in} \, \DualSet$) is denoted by $\pobj_{\param}(\x) \, {=} \, c_{\param}^{\top} \x$ (resp. $\dobj_{\param}(\y) \, {=} \, b_{\param}^{\top} \y$).
        The optimal value of $P_{\param}$ (resp. $D_{\param}$) is denoted by $\popt_{\param}$ (resp. $\psi^{\star}_{\param}$).
        Optimal solutions to the primal and dual problems are denoted by $\xopt_{\theta}$ and $\yopt_{\theta}$.
        By LP duality, the primal- and dual-optimal values are equal, and any primal (resp. dual) feasible solution yields a valid upper (resp. lower) bound on the optimal value, i.e.,
        \begin{align*}
            % \small
            \forall \param \, {\in} \, \ParamSet,
            \forall (\x, \y) \, {\in} \, \PrimalSet {\times} \DualSet,
            \ 
            \dobj_{\param}(\y) 
            \leq 
            \dopt
            = 
            \popt
            \leq
            \pobj_{\param}(\x).
        \end{align*}
        Finally, given $(\x, \y) \, {\in} \, \PrimalSet {\times} \DualSet$, define the \emph{duality gap}
        \begin{align}
            \label{eq:background:duality_gap}
            \Gap_{\theta}(\x, \y) = \pobj_{\param}(\x) - \dobj_{\param}(\y) \geq 0.
        \end{align}
        The duality gap provides a measure of how close to optimum a primal-dual solution is.
        Namely,
        \begin{subequations}
        \label{eq:background:duality_gap_suboptimality}
        \begin{align}
            \Gap_{\param}(\x, \y) &\geq \pobj_{\param}(\x) - \popt_{\param} \geq 0,\\
            \Gap_{\param}(\x, \y) &\geq \dopt_{\param} - \dobj_{\param}(\y) \geq 0.
        \end{align}
        \end{subequations}
        It is important to note that the duality gap only requires the knowledge of a primal-dual \emph{feasible} solution, i.e., \emph{$\Gamma_{\param}(\x, \y)$ can be computed without knowing the problem's optimal value}.
        
        % \noindent
        \emph{Remark:} 
            The LP setting is not a very restrictive assumption, as the proposed framework readily applies to nonlinear convex problems using conic duality.
            Readers are referred to \cite{ben2001lectures} for a more detailed review of conic optimization and duality. 
  
    \subsection{Hybrid Solver}
    \label{sec:hybrid_solver:framework}

        The paper assumes access to primal and dual optimization proxies that produce primal and dual solutions, respectively.
        Formally, let $\pproxy$ and $\dproxy$ denote the primal and dual proxy models respectively, and let $\xpred = \pproxy(\param)$ and $\ypred = \dproxy(\theta)$ denote the predicted primal and dual solutions for parameter $\param \in \ParamSet$. $\alpha$ (resp. $\beta$) refers to the primal (resp. dual) proxy's trainable parameters.
        It is further assumed that both proxies produce feasible solutions, i.e.,
        \begin{align}
            \label{eq:primal_dual_feasible_proxy}
            \forall \param \in \ParamSet,
            \left( \pproxy(\param), \dproxy(\param) \right)
            \in \mathcal{X}_{\param} \times \mathcal{Y}_{\theta}.
        \end{align}
        The assumption of primal-dual feasibility in Eq. \eqref{eq:primal_dual_feasible_proxy} is not too restrictive.
        Indeed, state-of-the-art neural network architectures such as $\Pi$net \cite{pinet} or RAYEN \cite{tordesillas2023rayen} can impose arbitrary convex constraints on their outputs.
        In the specific case of economic dispatch problems, dedicated repair layers have been proposed that ensure primal feasibility \cite{chen2023end}.
        In addition, \cite{tanneau2024dual} show how to build dual-feasible proxies for convex problems with bounded variables, which is the case for the economic dispatch problems considered in this work.
        Section \ref{sec:arch} presents the detailed architectures used for primal and dual proxies.

        \begin{algorithm}[!t]
           \caption{Hybrid solver inference}
           \label{algo:fallback}
            \textbf{Input}: query $\param$, optimality threshold $\threshold$
        
            \textbf{Output}: feasible, $\epsilon$-optimal prediction $(\tilde{\x},\,\tilde{\y})$\\[-1em]
        \begin{algorithmic}[1]
            \STATE Predict $(\xpred,\,\ypred)\in\PrimalSet{\times}\DualSet$
            \STATE Compute $\hat{g}=\Gap_\param(\hat{\mathbf{x}},\,\hat{\mathbf{y}})$
            \IF{$\hat{g} \leq \epsilon$}
            \RETURN $(\xpred,\,\ypred)$
            \ENDIF
            \RETURN $(\xopt_{\theta},\,\yopt_{\theta})$
        \end{algorithmic}
        \end{algorithm}

        The proposed hybrid solver is presented in Algorithm \ref{algo:fallback}.
        The solver takes as input the instance's parameter value $\param$ and a user-defined optimality tolerance $\epsilon$.
        First, a primal-dual solution $(\xpred, \ypred)$ is predicted using the primal and dual proxies, whose duality gap $\hat{g}$ is computed using \eqref{eq:background:duality_gap}.
        If the duality gap $\hat{g}$ is below the optimality tolerance $\epsilon$, the predicted solution $(\xpred, \ypred)$ is returned.
        Otherwise, an optimal solution $(\xopt_\theta, \yopt_\theta)$ is computed using an optimization solver.
        % Theorem \ref{thm:worst_case_performance} shows that the worst-case optimality gap is at most $\epsilon$.
        Thus, the worst-case optimality gap of the hybrid solver is at most $\epsilon$.

        % \begin{theorem}
        %     \label{thm:worst_case_performance}
        %     Let $\param \in \ParamSet$ and $\epsilon \geq 0$, and let $(\xsol, \ysol)$ be the solution returned by Algorithm \ref{algo:fallback} with input $\param, \epsilon$.
        %     The solution $(\xsol, \ysol)$ is at most $\epsilon$-suboptimal, i.e.,
        %     \begin{align*}
        %         \pobj_{\param}(\xsol) - \popt_{\param} \leq \epsilon
        %         \quad \text{and} 
        %         \quad
        %         \dopt_{\param} - \dobj_{\param}(\ysol) \leq \epsilon.
        %     \end{align*}
        % \end{theorem}
        % \begin{proof}
        %     Note that $(\xsol, \ysol)$ is always primal-dual feasible by construction.
        %     If $\hat{g} > \epsilon$, the result is immediate as $(\xsol, \ysol)$ is optimal.
        %     If $\hat{g} \leq \epsilon$, recall that $(\xpred, \ypred) \in \mathcal{X}_{\param} \times \mathcal{Y}_{\param}$.
        %     The result is a direct consequence of  \eqref{eq:background:duality_gap_suboptimality}.
        % \end{proof}
        
        It is important to note that the core enabler of the proposed framework is the availability of $\hat{g}$ as a self-contained certificate of sub-optimality, by leveraging LP duality and the feasibility of the predicted solution $(\xpred, \ypred)$.
        \emph{This key result makes it possible to detect ``bad" predictions online, without ground truth information, and call the optimization solver only when needed.}
        To the best of the authors' knowledge, this is the first data-driven hybrid framework that provides {\textit{controllable}} worst-case optimality guarantees without any offline verification step.

\section{Joint Primal-Dual Training}
\label{sec:primal_dual_training}

The standard self-supervised optimization proxy training performs
empirical risk minimization (maximization) using the primal (dual) objective value \cite{van2025optimization}. 
Using a finite set of samples $\Dataset$, the primal proxy training problem reads
\begin{equation}
    \begin{aligned}
    \min_{\pweights}\quad &\frac{1}{|\Dataset|}\, {\sum}_{\param\in\Dataset} \; 
     c_\param^\top\left(\pproxy(\param)\right)
    \end{aligned}
\end{equation}
Optimization proxies are typically relatively small (under 10M parameters), making it tractable to train 
the primal and dual models in parallel, on a single consumer GPU
even for large scale power systems\footnote{The largest case considered in the experiments, \texttt{9241\_pegase},
requires only 8GB of GPU memory for parallel training.}.
Training the primal and dual models in parallel
has several benefits, as discussed next.
The parallel empirical risk minimization problem reads
\begin{align}\label{eq:erm}
    \min_{\pweights,\dweights}\quad &\frac{1}{|\Dataset|}\, {\sum}_{\param\in\Dataset} \enspace 
     \Gap_\param\big(\pproxy(\param),\,\dproxy(\param)\big)
\end{align}
where $\pweights$ and $\dweights$ represent the trainable parameters of the neural networks
in the primal and dual optimization proxies respectively. Note that in terms of training gradients,
the problem is totally separable in $\pweights$ and $\dweights$. Furthermore, it recovers exactly
the primal (dual)-only training problems due to the definition of $\Gap_\param$.

{e
Although prior work has considered learning schemes that produce estimates of both primal and dual solutions, to the authors' knowledge, this work is the first to directly use the duality gap as the training loss. Enabled by the feasibility of the predictions, and in contrast to prior work based on alternating primal-dual updates \cite{park2023self,li2024pdhg} and penalty-based losses \cite{iftakher2025physics,arvind2024karush}, the proposed duality gap loss method yields a streamlined implementation by introducing no additional hyperparameters. Furthermore, the duality gap loss results in more stable and predictable training due to its separable form, since the primal (resp. dual) training gradient does not depend on the current performance of the dual (resp. primal) network. The streamlined and stabilized training is important for applications such as real-time risk analysis and market clearing where rapid automated re-training is required \cite{stover2022just}.
}

\subsection{Normalization}
A key feature of the primal-dual setting is that it allows for principled normalization.
Besides not relying on the optimal value, which is unknown at training and inference time, it is important that the  normalization retains the self-certifying property -- that the duality gap bounds from above the true optimality gaps
 -- when such a guarantee is needed. In these cases, e.g. to implement Algorithm \ref{algo:fallback} for normalized optimality thresholds $\epsilon\%$,
the predicted dual objective value is used, since it bounds from below the optimal value.
This yields the (proper) normalized gap
\begin{align}
\label{eq:normgap}
\bar{\Gap}_\param(\x, \y) \coloneq \frac{\Gap_\param(\x, \y)}{\dobj_{\param}(\y)}
\end{align}
which bounds from above the ground truth optimality gaps of the primal and dual predictions, i.e. replacing $\Gap_\param$ with $\bar{\Gap}_\param$ in Step 2 of Algorithm~\ref{algo:fallback} allows to certify
\begin{align*}
    \frac{\pobj_{\param}(\x) - \popt_{\param}}{\popt_{\param}} \leq \epsilon\%,\quad\text{and}\quad
    \frac{\dopt_{\param} - \dobj_{\param}(\y)}{\dopt_\param}  \leq \epsilon\%.
\end{align*}

In contexts where an exact bound on the true normalized optimality gap is not required, e.g. for normalizing the loss during training, estimates of the optimal value can be used.
In particular, the experiments in Section \ref{sec:results} normalize the training loss using the midpoint:
\begin{align}
\tilde{\Gap}_\param(\x, \y) \coloneq \frac{\Gap_\param(\x, \y)}{(\pobj_{\param}(\x)+\dobj_{\param}(\y))/2}
\end{align}
Note that the denominator is treated as a constant w.r.t. $\alpha$ and $\beta$
when performing automatic differentiation during training,
and validation metrics use the proper normalized gap \eqref{eq:normgap}.

\subsection{Targeting an optimality tolerance}

When jointly training primal-dual feasible optimization proxies
for later use in a hybrid solver, practitioners often already have
a target optimality threshold $\epsilon$ in mind.
In this case, the following ReLU/hinge loss more directly optimizes the speedup of the hybrid solver:
\begin{subequations}\label{eq:maxloss}\begin{align}
    \min_{\pweights,\dweights}\quad &\frac{1}{|\Dataset|}\, {\sum}_{\param\in\Dataset} \enspace {\Gap}^{\epsilon}_\param
     \\[0.1em]
     \text{where} \quad &
     {\Gap}^{\epsilon}_\param \coloneq\max(\Gap_\param\big(\pproxy(\param),\,\dproxy(\param)\big)-\epsilon, 0)
    \end{align}
\end{subequations}
since it focuses on improving the proxies at points $\param$
where $\Gap_\param\big(\pproxy(\param),\,\dproxy(\param)\big)>\epsilon$,
which would trigger the fallback to the classical solver in Algorithm \ref{algo:fallback}. The experiments compare $\Gap^0$ to $\Gap^{1\%}$; note that $\Gap^0$ recovers \eqref{eq:erm} due to \eqref{eq:background:duality_gap}.

\subsection{Sampling without replacement}
\label{sec:sampling}

Optimization proxies are typically trained
for thousands of epochs with $|\Dataset|$ in the tens to hundreds of thousands \cite{khaloie2024review},
where one epoch corresponds to one ``complete traversal'' through $\Dataset$;
in other words, each sample is seen many times during training. 
However, unlike the supervised learning setting where training data instances must be solved apriori --
representing a considerable computational burden --
in the self-supervised setting, generating new samples requires minimal computation. 
Thus, the paper proposes to \textit{resample $\Dataset$ at each epoch
so that each query is only seen once throughout training}.

This is equivalent to using a very large finite dataset, though can be implemented much more efficiently.
In particular, the need for storage and CPU-GPU transfer of training data batches is eliminated
since batches can be sampled on-demand, directly on the GPU.  
Although efficient and simple to implement, 
such implementations are uncommon due to the lack of a dependable convergence measure in the primal (dual)-only setting. 
Indeed, even self-supervised learning implementations are often based around finite datasets
in order to leverage pre-computed optimal solutions to implement features such as performance-based dynamic scheduling of learning rate, checkpointing, and termination, necessitating labeling at least the validation set samples and increasing the overall training time. Training the primal and dual proxies jointly enables implementing these crucial features using reliable metrics based on the (normalized) duality gap, fully eliminating the requirement for labeled data to begin training. Importantly, this enables immediate and trustworthy (re-)training within larger workflows.

\section{Numerical Experiments}
\label{sec:exp}
This section first describes the economic dispatch formulation used in this work
as well as the architectures used to implement the primal and dual proxies.
Then, experimental results are presented, using the hybrid solver framework
to accelerate solving batches of economic dispatch instances
on large-scale power system benchmarks 1354\_pegase, 2869\_pegase, and 9241\_pegase from PGLib \cite{pglib}.
Summary statistics for each benchmark are included in Table~\ref{tab:cases}.

\subsection{Economic Dispatch Formulation}
\label{sec:exp:formulation}

    Consider a power grid comprising $N$ buses, $E$ branches, $D$ loads and $G$ generators.
    The economic dispatch problem is formulated as the parametric LP problem
    \begin{subequations}
    \label{eq:ed primal_optimization}
        \begin{align}
\min_{\PG,\PF,\bm\xi} \quad 
                & c^{\top} \PG + M e^{\top} \bm\xi 
                \label{eq:ed:primal:obj}\\
            \text{s.t.} \quad
            & \label{eq:ed:primal:PTDF}
                \Phi A_{g} \PG - \PF = \Phi A_{d} \PD && [\bm\pi]\\
            & \label{eq:ed:primal:power_balance}
                e^{\top}\PG = e^{\top} \PD && [\bm\lambda]\\
            & \label{eq:ed:primal:pg_bounds}
                \underline{p} \leq \PG \leq \overline{p} && [\underline{\mathbf{z}}, \overline{\mathbf{z}}]\\
            & \label{eq:ed:primal:pf_bounds}
                \underline{f} - \bm\xi \leq \PF \leq \overline{f} + \bm\xi  && [\underline{\bm\mu},\overline{\bm\mu}] \\
            & \label{eq:ed:primal:positive_violations}
                \bm\xi \geq 0 && [\mathbf{y}] 
        \end{align}
    \end{subequations}
    parametrized by the vector of active power demands $\PD \, {\in} \, \mathbb{R}^{D}$.
    Decision variables $\PG \, {\in} \, \mathbb{R}^G$, $\PF\, {\in} \,\mathbb{R}^E$ and $\bm\xi \, {\in} \, \mathbb{R}^{E}$ denote active power dispatch, active power flow and line overflows, respectively.
    The notation $e$ is used to denote an appropriately-sized vector whose entries are all equal to $1$.
    The objective \eqref{eq:ed:primal:obj} minimizes active power generation costs and thermal violation penalties, where $M$ is a large positive constant, set to 150,000 in the experiments.
    Constraint \eqref{eq:ed:primal:PTDF} expresses power flows using the network's Power Transfer Distribution Factor (PTDF) matrix $\Phi$ and generator and load incidence matrices $A_{g}$ and $A_{d}$.
    Constraint \eqref{eq:ed:primal:power_balance} enforces global power balance.
    Constraints \eqref{eq:ed:primal:pg_bounds} and \eqref{eq:ed:primal:pf_bounds} enforce bounds on active power dispatch and power flows, respectively.
    Dual variables associated to each constraint are indicated between brackets.
    The dual of the economic dispatch reads
    \begin{subequations}
    \label{eq:dual_optimization-a}
    \begin{align}
        \max_{\substack{\bm\lambda, \bm\pi,  \underline{\mathbf{z}}, \overline{\mathbf{z}},\\\underline{\bm\mu}, \overline{\bm\mu},\mathbf{y}}} \quad
            & \hspace{-5em}\substack{\displaystyle \bm\lambda e^{\top}\PD
            + (\Phi A_{d} d)^{\top} \bm\pi
            \\\displaystyle \quad\quad\quad\hspace{2.3pt}\quad\quad\quad\quad\quad\quad+ \underline{f}^{\top} \underline{\bm\mu} - \overline{f}^{\top}\overline{\bm\mu} 
            + \underline{p}^{\top} \underline{\mathbf{z}} - \overline{p}^{\top} \overline{\mathbf{z}}}\\[-0.3em]
        \text{s.t.} \quad 
            & \bm\lambda e + (\Phi A_{g})^{\top} \bm\pi + \underline{\mathbf{z}} - \overline{\mathbf{z}} = c\\
            & -\bm\pi + \underline{\bm\mu} - \overline{\bm\mu} = 0\\
            & \underline{\bm\mu} + \overline{\bm\mu} + \mathbf{y} = Me\\
            & \underline{\bm\mu}, \overline{\bm\mu}, \underline{\mathbf{z}}, \overline{\mathbf{z}}, \mathbf{y} \geq 0
    \end{align}
    \end{subequations}

    To generate the training, validation, and testing samples,
    the PGLearn \cite[Algorithm 1 and Table 1]{klamkin2025pglearn} method is used to sample
    $\PD$ such that a wide range of total power demand is covered, including the challenging high-load regime.
    Note that unlike the PGLearn datasets, which include optimal solutions,
    as described in Section \ref{sec:sampling}
    the implementation in this paper does not collect solutions for training nor validation samples,
    allowing it to start training immediately.

    To implement the CPU solver baseline as well as the fallback in Algorithm \ref{algo:fallback}, a lazy constraint approach is used.
    Algorithm \ref{algo:lazy} describes the procedure, which is implemented in JuMP 1.28.0 \cite{jump} using the HiGHS 1.11.0 \cite{highs} solver.
    Using lazy constraints for the thermal limits drastically improves solve times; {for 1354\_pegase, Algorithm \ref{algo:lazy} is approximately 15x faster than an equivalent sparse phase-angle formulation, for 2869\_pegase 45x faster, and for 9241\_pegase 250x faster.}
Note that testing samples are solved individually offline to facilitate
reporting results for different levels of $\epsilon$, e.g. Figure \ref{fig:dist}. Furthermore, model building time is excluded from the solve time, and perfect sample-wise parallelism is emulated with 24 CPUs,
i.e., the solve time for a batch of $N$ samples each with solve times $t_i$ is computed
using the ideal makespan bound $\max(\frac{1}{24}\sum_it_i,\,\max_i(t_i))$.

\begin{algorithm}[t]
   \caption{Lazy PTDF solver}
   \label{algo:lazy}
    \textbf{Input}: query $\PD$

    \textbf{Output}: optimal solution $(\PG,\PF,\bm\xi)$\\[-1em]
\begin{algorithmic}[1]
    \STATE Initialize $\mathcal{C}\leftarrow\emptyset$
    \WHILE{true}
    \STATE Solve
    % \begin{align*}%
    \begin{equation*}%
    \begin{aligned}%
        \min_{\PG,\PF,\bm\xi} \quad & c^{\top} \PG
        + M e^{\top} \bm\xi 
        \\[-0.3em]
        \text{s.t.} \quad
        & (\Phi A_{g} \PG)_i - \PF_i = (\Phi A_{d} \PD)_i & \forall i\in\mathcal{C}\quad\\
        & \eqref{eq:ed:primal:power_balance}-\eqref{eq:ed:primal:positive_violations}%
    \end{aligned}%
    \end{equation*}%
    % \end{align*}%
    \STATE Compute $\PFtil\leftarrow \Phi A_{g} \PG-\Phi A_{d} \PD$
    \STATE Compute $\tilde{\bm\xi}=\max\big(\max(0, \PFtil-\overline{f}),\, \max(0,\underline{f}-\PFtil)\big)$
    \FOR{each branch $i$ in 1:$E$}
      \IF{$\tilde{\bm\xi}_i > 0$ \textbf{and} $i\notin\mathcal{C}$}
         \STATE $\mathcal{C}\leftarrow\mathcal{C}\cup\{i\}$ \quad\quad\quad\quad\COMMENT{add constraint $i$}
      \ENDIF
    \ENDFOR
    \IF{no constraints were added}
    \RETURN $(\PG,\PFtil,\tilde{\bm\xi})$
    \ENDIF
    \ENDWHILE
\end{algorithmic}
\end{algorithm}

\subsection{Feasibility Guarantees}
\label{sec:arch}

    In order to guarantee the feasibility of the primal predictions,
    the experiments implement the following inference procedure,
    based on the proportional response layer \cite{chen2023end}:
    \begin{enumerate}
        \item Predict \mbox{$\PGtil\in\{\PG\mid \underline{p}\leq {\PG}\leq \overline{p}\}$}
        \item Recover $\PGhat$ by using the proportional response layer \cite[Eq.~4]{chen2023end} to project $\tilde{\mathbf{p}}^\text{g}$ onto the hypersimplex \mbox{$\{\PG\mid e^\top {\PG} = e^\top \PD,\;\underline{p}\leq {\PG}\leq \overline{p}\}$}, i.e.
        \begin{subequations}
        \begin{align*}
        \PGhat\leftarrow\begin{cases}
            (1-\eta^\uparrow)\PGtil + \eta^\uparrow\overline{p} & \text{if} \enspace e^\top\PGtil< e^\top\PD \\
            (1-\eta^\downarrow)\PGtil + \eta^\downarrow\underline{p} & \text{if} \enspace e^\top\PGtil\geq e^\top\PD
        \end{cases}
        \end{align*}
        \begin{align*}
\text{where}\enspace\eta^\uparrow\coloneq\frac{e^\top\PD-e^\top\PGtil}{e^\top\overline{p}-e^\top\PGtil} \enspace\text{and}\enspace \eta^\downarrow\coloneq\frac{e^\top\PGtil-e^\top\PD}{e^\top\PGtil-e^\top\underline{p}} 
        \end{align*}
        \end{subequations}
        \item Recover $\PFhat{\,=\,}\Phi A_g \PGhat {\,-\,} \Phi A_d \PD$.
        \item Recover \mbox{$\hat{\bm\xi}=\max\big(\max(0, \PFhat-\overline{f}),\, \max(0,\underline{f}-\PFhat)\big)$.}
    \end{enumerate}

\begin{table*}[t]
    \centering
    \caption{Aggregated performance results across different weight initializations}
\begin{tabular}{ll|rrr|rrr}
\toprule
Case & Loss & 100x $\epsilon$\% & 500x $\epsilon$\% & 1000x $\epsilon$\% & $N$x $0.5\%$ & $N$x $1\%$ & $N$x $2\%$ \\
\midrule
1354\_pegase &  $\Gap^0$ & \textbf{0.43{\tiny$\pm$0.01}} & 0.96{\tiny$\pm$0.17} & 1.36{\tiny$\pm$0.20} & \textbf{168.17{\tiny$\pm$35.60}} & 588.97{\tiny$\pm$199.74} & 1646.03{\tiny$\pm$199.75} \\%[0.3em]
1354\_pegase &  $\Gap^{1\%}$ & 0.65{\tiny$\pm$0.07} & \textbf{0.75{\tiny$\pm$0.08}} & \textbf{0.80{\tiny$\pm$0.09}} & 1.90{\tiny$\pm$0.54} & \textbf{1782.84{\tiny$\pm$297.92}} & \textbf{2163.81{\tiny$\pm$2.29}} \\%[0.05em]
\midrule
2869\_pegase &  $\Gap^0$ & \textbf{0.47{\tiny$\pm$0.03}} & 1.17{\tiny$\pm$0.05} & 1.49{\tiny$\pm$0.05} & \textbf{109.01{\tiny$\pm$6.99}} & 332.89{\tiny$\pm$26.66} & 1672.71{\tiny$\pm$119.89} \\%[0.3em]
2869\_pegase &  $\Gap^{1\%}$ & 0.56{\tiny$\pm$0.04} & \textbf{1.10{\tiny$\pm$0.03}} & \textbf{1.36{\tiny$\pm$0.03}} & 50.27{\tiny$\pm$34.33} & \textbf{388.00{\tiny$\pm$32.21}} & \textbf{1798.52{\tiny$\pm$19.63}} \\%[0.05em]
\midrule
9241\_pegase &  $\Gap^0$ & \textbf{0.53{\tiny$\pm$0.03}} & 0.87{\tiny$\pm$0.02} & 2.88{\tiny$\pm$0.07} & \textbf{82.33{\tiny$\pm$18.62}} & 707.22{\tiny$\pm$57.06} & 958.89{\tiny$\pm$18.36} \\%[0.3em]
9241\_pegase &  $\Gap^{1\%}$ & 0.57{\tiny$\pm$0.03} & \textbf{0.85{\tiny$\pm$0.04}} & \textbf{1.18{\tiny$\pm$0.04}} & 64.04{\tiny$\pm$13.71} & \textbf{925.23{\tiny$\pm$39.92}} & \textbf{1113.27{\tiny$\pm$1.30}} \\%[0.05em]
\bottomrule
\end{tabular}
    \label{tab:main}
\end{table*}

    To guarantee dual-feasiblility, the paper adopts Smoothed Self-Supervised Learning (S3L) \cite{klamkin2024dual},
    which for the economic dispatch reads:
    \begin{enumerate}
        \item Predict $\hat{\bm\lambda}\in\mathbb{R}$ and $\hat{\bm\pi}\in[-M,M]^E$
        \item Recover $\hat{\underline{\bm\mu}}=\frac{\mu}{\overline{f}-\underline{f}}+\frac{\hat{\bm\pi}}{2} + \sqrt{(\frac{\mu}{\overline{f}-\underline{f}})^2+(\frac{\hat{\bm\pi}}{2})^2}$
        \item Recover $\hat{\overline{\bm\mu}}=\frac{\mu}{\overline{f}-\underline{f}}-\frac{\hat{\bm\pi}}{2} + \sqrt{(\frac{\mu}{\overline{f}-\underline{f}})^2+(\frac{\hat{\bm\pi}}{2})^2}$
        \item Compute $z=c - \bm\hat{\bm\lambda} e - (\Phi A_{g})^{\top} \bm\hat{\bm\pi}$
        \item Recover $\hat{\underline{\mathbf{z}}}=\frac{\mu}{\overline{\PG}-\underline{\PG}}+\frac{z}{2} + \sqrt{(\frac{\mu}{\overline{\PG}-\underline{\PG}})^2+(\frac{z}{2})^2}$
        \item Recover $\hat{\overline{\mathbf{z}}}=\frac{\mu}{\overline{\PG}-\underline{\PG}}-\frac{z}{2} + \sqrt{(\frac{\mu}{\overline{\PG}-\underline{\PG}})^2+(\frac{z}{2})^2}$
    \end{enumerate}
    
    Note that this procedure is only used during training.
    During inference, Dual Lagrangian Learning (DLL) \cite{tanneau2024dual} is used since
    it guarantees optimal completion given fixed $\hat{\bm\lambda}$ and $\hat{\bm\pi}$:
    \begin{enumerate}
        \item Predict $\hat{\bm\lambda}\in\mathbb{R}$ and $\hat{\bm\pi}\in[-M,M]^E$
        \item Recover $\hat{\underline{\bm\mu}}= \max(0, \hat{\bm\pi})$.
        \item Recover $\hat{\overline{\bm\mu}} = \max(0, -\hat{\bm\pi})$.
        \item Recover $\hat{\underline{\mathbf{z}}} = \max(0, c - \bm\hat{\bm\lambda} e - (\Phi A_{g})^{\top} \bm\hat{\bm\pi})$.
        \item Recover \mbox{$\hat{\overline{\mathbf{z}}} = \max(0, (\Phi A_{g})^{\top} \bm\hat{\bm\pi} + \bm\hat{\bm\lambda} e - c)$.}
    \end{enumerate}

Note that both the primal and dual recovery procedures require bounded predictions as part of Step 1;
the experiments use the ``double-softplus'' function to enforce the
element-wise bound constraints $l\leq x\leq u$ on the neural network predictions:
$\tilde{x}\coloneq\ln(1+e^{x-l})-\ln(1+e^{x-u})+l$.

Finally, note that although the hybrid solver framework is applicable for any architecture
which produces primal-dual feasible predictions, its performance depends on the inference time of the models,
including any repair steps. The procedures described above are chosen
due to their computational efficiency, enabling timely training and fast inference.

\subsection{Results}
\label{sec:results}

\begin{table}[!ht]
  \caption{Case sizes, timing statistics, and parameter counts}
  \label{tab:cases}
  \setlength{\tabcolsep}{4.5pt}
  \centering
  \begin{tabular}{l|rrr|rrr|r}
    \toprule
    Case name              & $|\PD|$ & $|\PG|$ & $|\PF|$ & $T^\text{Opt}_\text{240k}$ & $T^\text{ML}_\text{240k}$ & $T^\text{ML}_\text{Train}$ & {$|\alpha|{+}|\beta|$}\\
    \midrule
    {1354\_pegase}     &    673     &      260        &     1991 & 90.8s & 0.03s &28min & {1.3M}\\
    {2869\_pegase}      &    1491    &      510        &     4582 & 199.7s & 0.09s &32min & {2.5M} \\
    {9241\_pegase}    &    4895    &      1445       &     16049 & 711.2s & 0.64s & 70min & {7.4M} \\
    \bottomrule
  \end{tabular}
\end{table}

The experiments compare the hybrid solver to the parallelized classical solver described in Section \ref{sec:exp:formulation}.
For each case, two types of hybrid solvers are evaluated,
differing only in the loss function used to train the underlying proxy models:
$\Gap^0$ refers to the standard ERM-based loss \eqref{eq:erm} and
$\Gap^{1\%}$ refers to the ``ReLU'' loss \eqref{eq:maxloss} with a target optimality threshold of $\epsilon{=}1\%$.
All other hyperparameters are kept constant: proxies are trained using the
Adam optimizer with batch size 1024 and learning rate $10^{-3}{\rightarrow}10^{-5}$,
scheduled using the normalized mean duality gap on the validation set.
Specifically, the learning rate is multiplied by $0.95$
when the gap does not decrease by at least 0.01\% for 50 consecutive epochs.
The primal and dual proxies each use 4 hidden layers with dimension 256,
learned batch-normalization, and softplus activations.
Training is run for 5000 epochs with a (re-sampled, as discussed in Section~\ref{sec:sampling})
dataset size of 20,480. Validation is run every epoch using the same 10,240 (unlabeled) samples, and
checkpoints are saved based on the best mean normalized duality gap on the validation set.
Results are presented on a batch of 240,000 unseen testing samples.
This batch size is chosen to mimic performing a {1-day hourly-granularity simulation with 10,000 scenarios}.

The training and evaluation is run on an NVIDIA H200 GPU, implemented using PyTorch 2.8.0 \cite{torch}.
GPU inference uses \texttt{torch.compile} %\footnote{Default settings except for \texttt{dynamic=False}.}
including the proxy inference, repair, and objective value calculation.
This straightforward inference setup is used to keep the implementation simple;
further speedup can likely be achieved by improving this aspect.
Furthermore, time spent on data movement (inputs to GPU and outputs to CPU)
is \textit{included} in the inference time measurements.

Table \ref{tab:cases} includes the column $T^\text{Opt}_\text{240k}$,
reporting the solve time of the perfectly-parallelized classical solver for the test set.
The column $T^\text{ML}_\text{240k}$ reports the inference time of the proxy models;
across all cases, the ML models return predictions for all 240,000 samples in milliseconds,
far exceeding the throughput of the parallelized classical solver.
The column $T^\text{ML}_\text{Train}$ shows the training time for the ML models;
about 30 minutes for 1354\_pegase and 2869\_pegase
and just over an hour for 9241\_pegase.
Finally, the $|\alpha|{+}|\beta|$ column reports the total number of trainable parameters across both the primal and dual models; the largest case uses under 7.5M parameters.

The ``$N$x $\epsilon$\%'' metric is used to analyze the performance of the proposed hybrid solver; it
captures at what percent optimality tolerance $\epsilon$\% the hybrid solver achieves an $N$x speedup. In other words,
Pareto-optimal hybrid solvers achieve the largest $N$x speedups at the lowest $\epsilon$\% tolerances.
Note that, in contrast to NNV-based approaches which only provide a global worst-case bound, the hybrid solver allows practitioners to choose
the target optimality threshold $\epsilon$\%, then evaluate the speedup.
Table \ref{tab:main} includes such evaluations for $\epsilon{=}\{0.5\%,1\%,2\%\}$,
as well as ``inverse'' results showing at what $\epsilon\%$ level an $N{=}\{100,500,1000\}$-times speedup is achieved.
Furthermore, Figure \ref{fig:dist} shows $N$ as a function of $\epsilon$.

Table \ref{tab:main} shows the hybrid solver performance across cases and when using
different training losses for the underlying proxies. Notably, all configurations achieve
substantial speedups at practical optimality tolerance levels, all achieving 1000x speedup
below 3\% tolerance. When using the $\Gap^{1\%}$ loss configurations, 100x is achieved
below $0.7\%$ across all cases and 1000x is achieved below $1.4\%$ for all cases. 
Furthermore, the difference between the $\Gap^0$ and $\Gap^{1\%}$ rows highlights
the tradeoff associated with targeting an optimality tolerance;
speedup for tolerance levels below the target is traded for speedup at and above the target. Notably, $\Gap^{1\%}$ consistently outperforms $\Gap^0$ at the target tolerance (corresponding to the $N$x 1\% column), achieving speedups over 1500x for 1354\_pegase, over 350x for 2869\_pegase, and over 900x for 9241\_pegase. With a higher 2\% threshold, speedup is over 1000x for all cases; above 2000x for 1354\_pegase.

Figure \ref{fig:dist} further explores the impact of using a non-zero target threshold,
comparing the speedup-optimality tradeoff curves when using different underlying proxy models,
one trained using $\Gap^{0}$, labeled ERM, and the other using $\Gap^{1\%}$, labeled ReLU.
The figure shows that, at the 1\% target tolerance and above, using the $\Gap^{1\%}$ models results in larger speedups.
At optimality tolerances below the target, $\Gap^0$ begins to outperform $\Gap^{1\%}$. While $\Gap^{1\%}$ only achieves 10x speedup for thresholds below 0.6\%, $\Gap^0$ achieves over 100x speedup while guaranteeing at most 0.5\% sub-optimality.

\begin{figure}[t]
    \centering
    \includegraphics[width=0.99\linewidth]{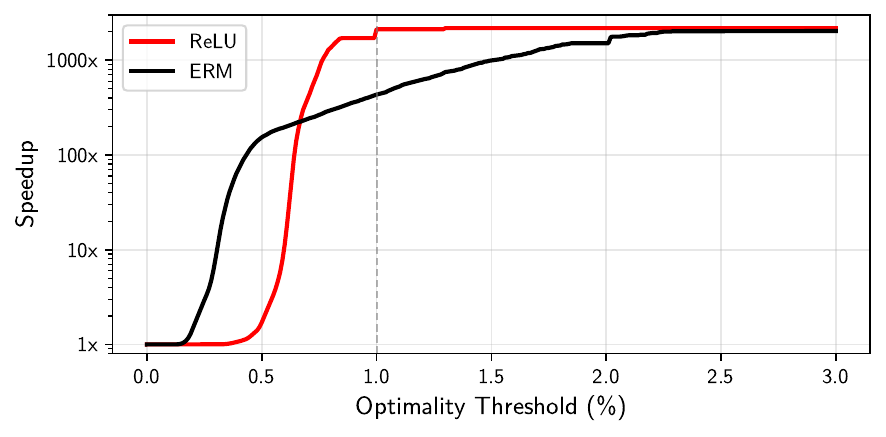}
    \caption{Speedup of hybrid solver on 1354\_pegase as a function of the optimality threshold $\epsilon$.}%, for neural network initialization seed 1.}
    \label{fig:dist}
\end{figure}

\section{Conclusion}
\label{sec:conclusion}

This paper has proposed a novel hybrid framework that accelerates large-scale computations in power systems operations and planning by integrating data-driven machine learning proxies and classical optimization solvers in a principled way.
The proposed hybrid solver offers substantial speedups \emph{and} worst-case performance guarantees, thus synergizing several recent developments in machine learning for power systems.
This approach further closes the gap between theoretical advances and the tight requirements of practical deployments, enabling trustworthy proxy-based market clearing with performance guarantees and real-time risk analysis for large-scale power systems.

Unlike neural network verification-based approaches, the proposed methodology does not impose strong restrictions on activation functions, does not require any post-training analysis, and provides \emph{adaptable} worst-case optimality guarantees.
The paper has conducted extensive computational experiments on large-scale power grids, corresponding to economic dispatch problems with over $10^5$ decision variables and constraints, and over $10^4$ parameters.
Numerical results demonstrate the scalability of the approach, achieving speedups on the order of 1000x over optimized classical solvers. 

% \balance

% \clearpage
% \section*{AI Usage Disclosure}

% \noindent No AI tools were used in the development of this paper.

\section*{Acknowledgments}
{
This material is based upon work supported by the National Science Foundation under Grant No. 2112533 and Grant No. DGE-2039655, and partially funded by Los Alamos National Laboratory’s Directed Research and Development project, “Artificial Intelligence for Mission (ArtIMis).” Any opinions, findings, and conclusions or recommendations expressed in this material are those of the author(s) and do not necessarily reflect the views of the sponsors.
}
% \balance
% \clearpage
\bibliographystyle{ieeetr}
\bibliography{bibliography}
% \balance

\endgroup
\end{document}